\documentclass{ieeeaccess}
\usepackage{cite}
\usepackage{amsmath,amssymb,amsfonts}
\usepackage{algorithmic}
\usepackage{graphicx}
\usepackage{textcomp}
\usepackage{booktabs}
\usepackage{multirow}
\usepackage{caption}

\def\BibTeX{{\rm B\kern-.05em{\sc i\kern-.025em b}\kern-.08em
    T\kern-.1667em\lower.7ex\hbox{E}\kern-.125emX}}
\begin{document}
\history{Date of publication xxxx 00, 0000, date of current version xxxx 00, 0000.}
\doi{10.1109/ACCESS.2017.DOI}

\title{MSA\textsuperscript{2}-Net: Utilizing Self-Adaptive Convolution Module to Extract Multi-Scale Information in Medical Image Segmentation }
\author{\uppercase{Chao Deng }\authorrefmark{1},
\uppercase{Xiaosen Li \authorrefmark{2}, and Xiao Qin,}.\authorrefmark{1},}
\address[1]{School of Artificial Intelligence, Nanning Normal University, Nanning, Guangxi, 530100, People's Republic of China (e-mail: chaodeng987@email.nnnu.edu.cn(Chao Deng); 110027@nnnu.edu.cn(Xiao Qin))}
\address[2]{School of Artificial Intelligence, Guangxi Minzu University, Nanning, Guangxi, 530300, People’s Republic of China (e-mail: xiaosenforai@gmail.com(Xiaosen Li))}

\tfootnote{This study was financially supported by STI2030-Major Projects (2021ZD0201900),Guangxi Key R\&D Project (Grant no.AA22068057).}

\markboth
{Author \headeretal: Preparation of Papers for IEEE TRANSACTIONS and JOURNALS}
{Author \headeretal: Preparation of Papers for IEEE TRANSACTIONS and JOURNALS}

\corresp{Corresponding author: Xiao Qin (e-mail: 110027@nnnu.edu.cn).}

\begin{abstract}
The nnUNet segmentation framework adeptly adjusts most hyperparameters in training scripts automatically, but it overlooks the tuning of internal hyperparameters within the segmentation network itself, which constrains the model's ability to generalize. Addressing this limitation, this study presents a novel Self-Adaptive Convolution Module that dynamically adjusts the size of the convolution kernels depending on the unique fingerprints of different datasets. This adjustment enables the MSA\textsuperscript{2}-Net, when equipped with this module, to proficiently capture both global and local features within the feature maps. Self-Adaptive Convolution Module is strategically integrated into two key components of the MSA\textsuperscript{2}-Net: the Multi-Scale Convolution Bridge and the Multi-Scale Amalgamation Decoder. In the MSConvBridge, the module enhances the ability to refine outputs from various stages of the CSWin Transformer during the skip connections, effectively eliminating redundant data that could potentially impair the decoder's performance. Simultaneously, the MSADecoder, utilizing the module, excels in capturing detailed information of organs varying in size during the decoding phase. This capability ensures that the decoder's output closely reproduces the intricate details within the feature maps, thus yielding highly accurate segmentation images. MSA\textsuperscript{2}-Net, bolstered by this advanced architecture, has demonstrated exceptional performance, achieving Dice coefficient scores of 86.49\%, 92.56\%, 93.37\%, and 92.98\% on the Synapse, ACDC, Kvasir, and Skin Lesion Segmentation (ISIC2017) datasets, respectively. This underscores MSA\textsuperscript{2}-Net's robustness and precision in medical image segmentation tasks across various datasets.
\end{abstract}

\begin{keywords}
Convolution, Medical Image Segmentation, Quartile statistics, Transformer
\end{keywords}

\titlepgskip=-15pt

\maketitle

\section{Introduction}
\label{sec:introduction}
\PARstart{M}{edical} image segmentation occupies a crucial role in medical diagnostics by enabling physicians to swiftly pinpoint pathogens\cite{RN1,RN2,RN3}. This process involves extracting and delineating specific anatomical structures or pathological lesions from varied medical imaging modalities, including computed tomography (CT), magnetic resonance imaging (MRI), and X-rays. Essentially, the aim of medical image segmentation is to assign a specific classification to each pixel or voxel in the image, which could represent distinct organs, tissues, lesions, and other relevant anatomical features.
Since the emergence of Convolutional Neural Networks (CNNs)\cite{RN5}, they have been widely applied in various image analysis tasks and have gradually become the dominant technological framework in medical image semantic segmentation. They are capable of efficiently extracting edges, textures, and structural information of organs through the mechanisms of local receptive fields and parameter sharing, making them particularly suitable for medical image processing scenarios with strong spatial local features. U-Net, as a typical representative, is the most popular solution in medical segmentation tasks. The U-Net architecture consists of an encoder, skip connections, and a decoder. The encoder conducts feature extraction through convolution operations and downsamples the feature maps; the decoder is responsible for restoring the detail of the feature maps through upsampling operations; to complete the feature map details lost during the feature extraction by the encoder, U-Net introduces skip connections between the encoder and decoder. U-Net has demonstrated exceptional performance in semantic segmentation tasks, significantly advancing the precision and generalization capabilities of these applications.
However, CNN is fundamentally a method focusing on local modeling\cite{RN6,RN7,RN8}. It is limited by the size of the convolutional kernels, which makes it challenging to capture multi-scale information in medical images. This limitation is particularly evident when dealing with multi-organ semantic segmentation tasks. Such constraints weaken the model's ability to integrate global contextual information, thereby affecting segmentation accuracy. To enhance the model's capability to capture multi-scale information, researchers have proposed various improvement strategies. For instance, MISSFormer\cite{RN28} introduces an Enhanced Transformer Context Bridge in different stages of the encoder outputs to capture multi-scale information; Dilated\cite{RN39} Convolution enlarges the receptive field by increasing the dilation rate, effectively capturing large-scale context information without the need for significantly increasing the number of parameters or computational complexity; Spatial Transformer Networks (STN)\cite{RN42} incorporate a learnable geometric transformation module, allowing the model to adaptively adjust spatial representations based on input features, thus enhancing its structural perception capabilities; Adaptive Spatial Feature Fusion (ASFF)\cite{fu2021adaptive} strategy proposes the use of a spatial filtering mechanism to integrate features of different scales, effectively mitigating information conflicts between different resolutions, and enhancing the consistency of semantic representations. Although these methods have improved CNN's ability to process multi-scale information to some extent, the size of convolutional kernels—a hyperparameter determining the model's receptive field—usually needs to be preset and cannot dynamically adjust according to the dataset trained on, which limits the model's generalization capabilities.
nnUNet\cite{RN33} adopts a method based on data fingerprints and pipeline fingerprints to represent key attributes of datasets and networks, enabling it to automatically adjust hyperparameters without human intervention. However, the hyperparameters adjusted by nnUNet are mainly focused on dataset processing aspects (such as voxel size, spacing, and batch size), and do not fully consider the adjustments of hyperparameters within the network structure (such as convolutional kernel size and padding). To address this challenge, this study introduces a Multi-Scale Adaptive Attention Network (MSA\textsuperscript{2}-Net), aimed at achieving precise segmentation of multi-organ medical images. MSA\textsuperscript{2}-Net employs an encoder-decoder architecture, with the Cross-Shape Window Transformer (CSWin)\cite{RN22} at the core of its encoder, utilizing a cross-window mechanism to effectively jointly simulate the local details and global semantics of the segmentation targets; in the skip connections, a Multi-Scale Convolution Bridge (MSConvBridge) is designed to alleviate the semantic gap between the encoder and decoder, enhancing the consistency and discriminability of the features; for the decoder part, a Multi-Scale Adaptive Decoder (MSADecoder) is adopted, which enhances the model's capability to capture multi-scale information through recursively nested adaptive convolution groups. The main contributions of this study can be summarized as follows:
\begin{enumerate}
\item Proposed adaptive convolution module that generates corresponding data fingerprints for different segmentation datasets and automatically adjusts the receptive field of convolutions based on these fingerprints, enhancing the model's generalizability.
\item An innovative Multi-Scale Convolution Bridge (MSConvBridge) is proposed based on adaptive convolution modules. This architecture utilizes a combination of multi-scale adaptive convolutions to delicately process the feature maps output by CSWin. Through this multi-layer approach, the MSConvBridge efficiently eliminates redundant striated textures between image features while maximally preserving the information within the feature maps, thereby significantly enhancing overall semantic consistency. This method not only optimizes feature representation but also strengthens the model's capability to handle complex scenes.
\item Multi-Scale Amalgamation Decoder (MSADecoder) employs recursive nested adaptive convolution groups, enabling the decoder to replenish information about small organs in the image while upscaling, and simultaneously preserving information about large organs in the image.
\end{enumerate}

\section{RELATE WORK}
\subsection{nnUnet}
In medical image segmentation networks, there are a plethora of hyperparameters. Researchers often have to adjust these hyperparameters through repeated experiments during the network design process. This adjustment is typically reliant on the individual researcher's experience, making the process highly inefficient. Excessive manual adjustments to the network structure can lead to overfitting for specific datasets. The impact of non-structural aspects of the network might have a greater influence on the segmentation task. Therefore, nnUNet was developed, focusing not on modifying the specific network architecture but on adjusting the hyperparameters in dataset preprocessing, training scripts, and post-processing. Although nnUNet's approach has improved training efficiency, it overlooks adjustments of the hyperparameters within the network structure itself. The adaptive convolution module proposed in this article builds on the nnUNet framework, capable of adapting the size of the convolutional receptive field according to the characteristics of the dataset.
\subsection{U-Net-like Semantic Segmentation Networks}
U-Net is a classical fully convolutional network architecture widely used for semantic segmentation tasks, comprising three key components: an encoder, a decoder, and skip connections. The architecture features a symmetrical U-shape, the left-side encoder path progressively extracts image features through successive downsampling layers, capturing higher-level semantic information while reducing spatial resolution; concurrently, the right-side decoder path reconstructs spatial resolution through upsampling operations to produce pixel-level predictions. To maintain spatial precision and mitigate semantic information loss, U-Net employs skip connections that directly transfer high-resolution features from shallow encoder layers to their corresponding decoder layers, a design that substantially enhances segmentation accuracy for boundary regions and small targets.
In Unet++\cite{RN26}, the encoder extracts multi-level semantic features through convolution and downsampling operations, generating multi-scale feature maps, while introducing dense skip connections to enhance the fusion efficiency of features at different semantic levels, and adopts a deep supervision mechanism during the decoding stage. With the continuous improvement in the resolution and diversity of remote sensing images, U-Net and its variants have been widely applied to medical image semantic segmentation tasks, making continuous progress in multi-scale feature extraction, boundary modeling, and contextual fusion. For example: Wu et al. proposed the High-Resolution Network (HRNet), which constructs multi-scale feature representations through multi-branch parallel convolutions and designs an adaptive spatial pooling module to enhance local contextual modeling capabilities; Majoli et al. integrated an edge detection module into the segmentation network to explicitly capture boundary features and improve organ contour recognition accuracy; Qi et al. proposed a spatial information reasoning structure that combines recurrent neural networks with 3D convolutions to enhance the network's spatial understanding ability and significantly improve organ detection performance\cite{qi2018age}.
Although existing methods have made many advances in the field of semantic segmentation of remote sensing images, they still face challenges such as complex network structures, high computational costs, insufficient boundary detail recovery capability in decoders, and poor generalization on medical image datasets. Current research mainly focuses on three core issues:
\begin{enumerate}
\item Accuracy-Complexity Tradeoff: Reducing model complexity while maintaining segmentation accuracy
\item Enhanced Detail Recovery: Improving fine-grained reconstruction capability for boundaries and small targets
\item Dataset Adaptability Optimization: Strengthening generalization adaptability to complex spatial features in medical imaging
\end{enumerate}

\section{METHOD}
\subsection{Self-Adaptive Convolution Module}
\begin{figure}
    \centering
    \includegraphics[width=1\linewidth]{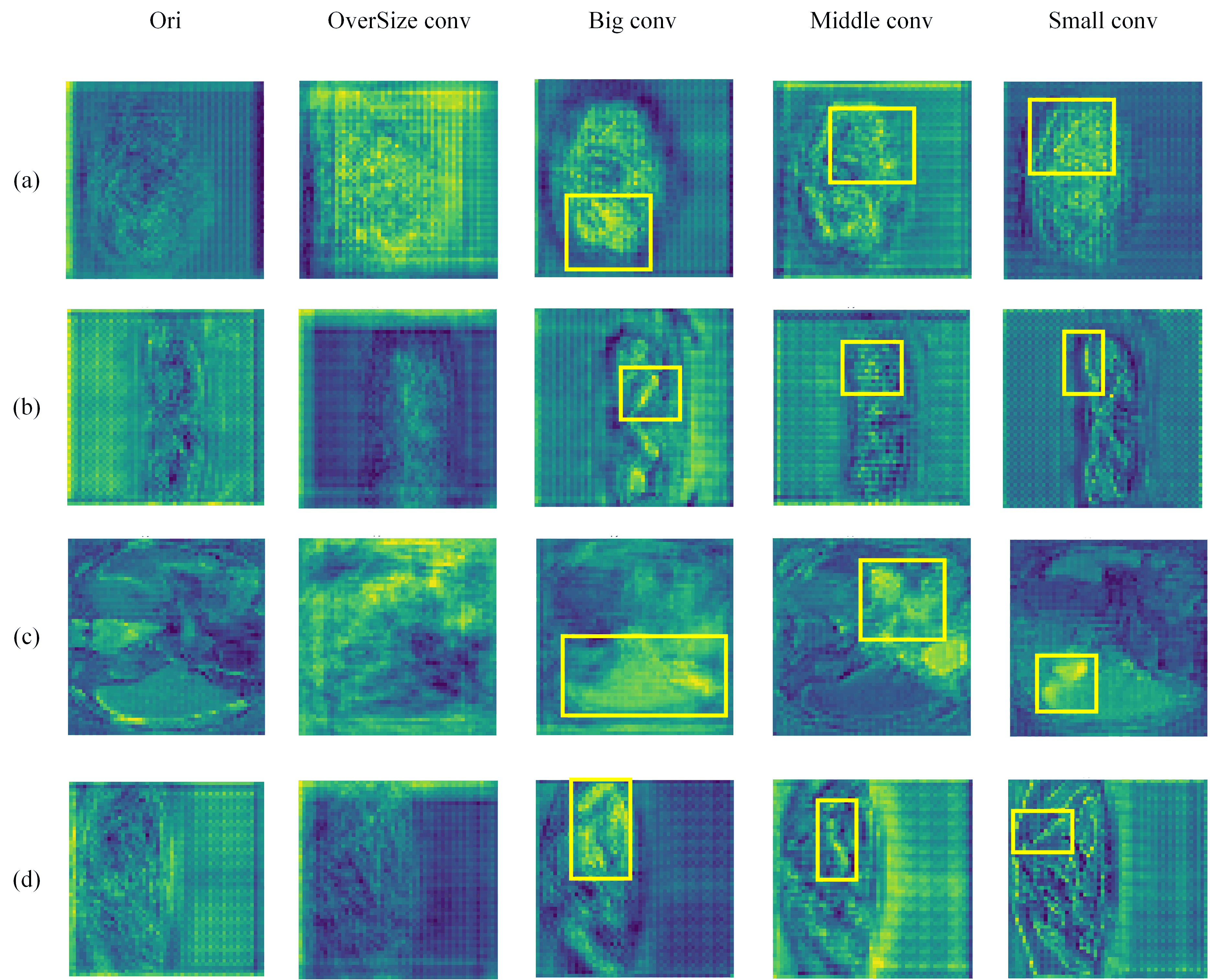}
    \caption{Impact of Different Kernel Sizes on Feature Maps }
    \label{fig:1}
\end{figure}

The convolutional operation is constrained by its receptive field, typically only capturing object information that matches its field size. As shown in Figure \ref{fig:1}(b), feature maps processed with large kernels amplify large organs while losing small organ details, whereas small convolution kernels exclusively highlight small organ features. In medical imaging, multiple visceral organs often interweave at the image center, surrounded by blank background regions. Therefore, kernel sizes must be adaptively designed according to target organ dimensions to prevent background interference. Figure \ref{fig:1}(c) demonstrates how over size convolution kernel sizes cause background contamination of feature maps, resulting in chaotic organ structure representation.

Self-Adaptive Convolution Module dynamically adjusts the receptive field of convolution based on the size of valid information within the feature maps, achieving efficient multi-scale information capture. The convolution kernel selection strategy for the adaptive convolution module is expressed by the following formula:
\begin{equation}
K_i=w_{\hat{i}},w_{\hat{i}}\in W_C
\end{equation}

$K_i$ denotes the convolutional kernel for the $i$-th convolutional operation, $W_C$ represents the candidate parameter matrix from which the convolutional kernel is selected, and $w_{\hat{i}}$ indicates the element in $W_C$ with index $\hat{i}$. The definition of the subscript $\hat{i}$ is given below:

\begin{equation}
\hat{i}=max\left(W_s\right),W_s\in\mathbb{R}^{4\times4}
\end{equation}
$W_s$ denotes the selection probability matrix, where each element in $W_s$ represents the selection probability of the corresponding element in the candidate parameter matrix $W_C$. $W_s$ is updated during each backpropagation iteration. The construction method for the candidate parameter matrix $W_C$ is presented below:
\begin{equation}
W_C=W_B\cdot W_{QS},W_B\in\mathbb{R}^{4\times1},W_{QS}\in\mathbb{R}^{1\times4}
\end{equation}
$W_B$ represents the baseline kernel matrix, following the configuration in TransXNet, with its elements set to $[1, 3, 5, 7]$. $W_{QS}$ denotes the quartile shift matrix, whose construction method is described below:
\begin{equation}
W_{QS}=1+QS\left(Dataset\right)
\end{equation}
Where $QS$ denotes the quartile statistical operation, and Dataset represents the currently processed dataset in MSA\textsuperscript{2}-Net. $W_B^T$ is the transpose matrix of $W_B$. Self-Adaptive Convolution Module dynamically adjusts the convolutional receptive field according to dataset characteristics, ensuring the kernel size is neither too large nor too small - precisely covering most sub-features in the feature maps to achieve efficient multi-scale information extraction.
\begin{figure*}
    \centering
    \includegraphics[width=1\linewidth]{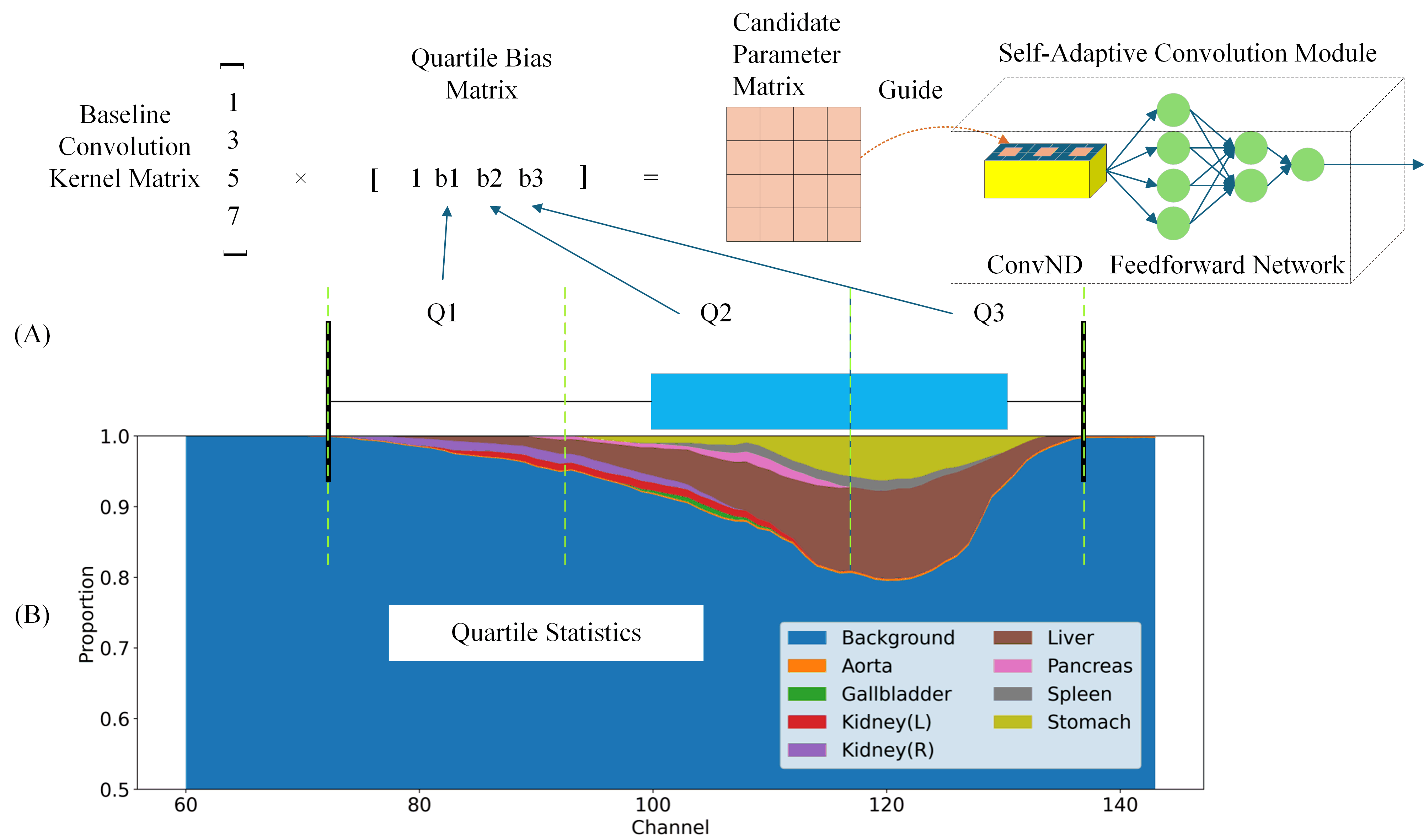}
    \caption{(A) Schematic diagram of the Self-Adaptive Convolution Module and its parameter auto-update workflow. (B) Area proportion distribution of different organs in feature maps from the Synapse dataset. The x-axis represents the variation in channel dimensions (starting from channel 60 as the first 60 channels contain only background information). The y-axis represents the area proportion of different organs in the feature maps at each channel dimension.}
    \label{fig:2}
\end{figure*}
\subsection{MSA\textsuperscript{2}-Net Overall Structure}
\begin{figure*}
    \centering
    \includegraphics[width=1\linewidth]{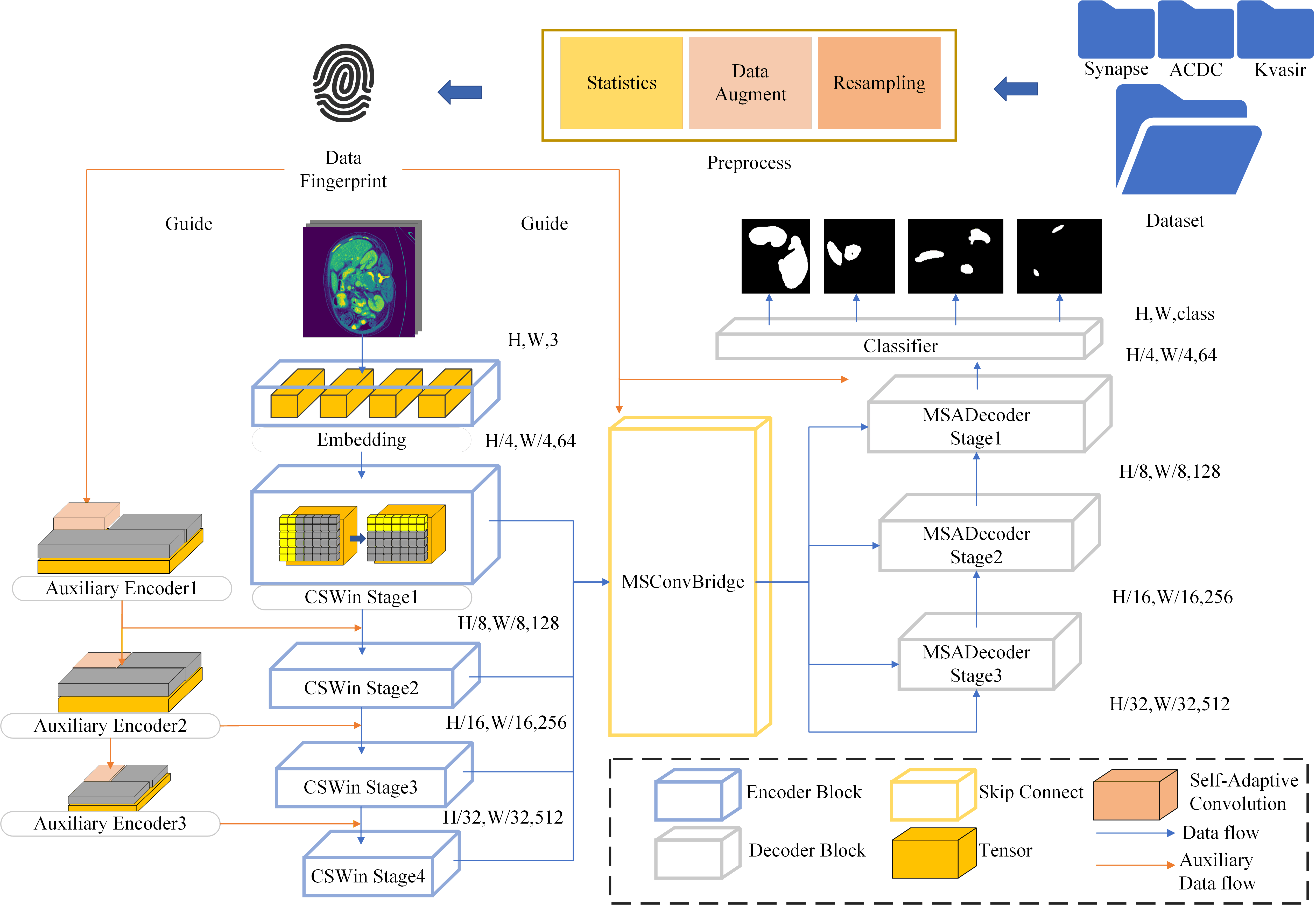}
    \caption{Architectural Overview of MSA\textsuperscript{2}-Net.}
    \label{fig:3}
\end{figure*}

The specific architecture of MSA\textsuperscript{2}-Net is illustrated in Figure \ref{fig:3}. The left portion of Figure \ref{fig:3} shows the encoder component of MSA\textsuperscript{2}-Net, while the right portion displays the Multi-Scale Adaptive Decoder (MSADecoder). The middle section represents the Multi-Scale Convolution Bridge (MSConvBridge).
For a given image $x\in\mathbb{R}^{H\times W\times3}$, the encoder outputs multi-scale feature maps at different stages:
\begin{equation}
F=Encoder\left(x\right)
\end{equation}
$F=[F_1,F_2,F_3,F_4]$ represents the feature maps from each encoding stage, where spatial resolution progressively decreases while channel depth gradually increases. These features $F$ are then fed into the MSConvBridge for refinement:
\begin{equation}
\hat{F}=MSConvBridge\left(F\right)
\end{equation}
Finally, MSADecoder generates the segmentation map output by aggregating and upsampling these refined features:
\begin{equation}
S=MSADecoder\left(\hat{F}\right),S\in\mathbb{R}^{H\times W\times C}
\end{equation}
where $C$ denotes the number of categories.
\subsection{Encoder}
To better adapt CSWin for medical image segmentation tasks, we modified the original encoder design of CSWin to preserve detailed information in feature maps while performing downsampling for feature extraction. MSA\textsuperscript{2}-Net integrates the CSWin encoder with ResNet, where the CSWin encoder serves as the primary encoder and the ResNet equipped with adaptive convolution modules acts as the auxiliary encoder. The internal workflow at each encoding stage can be expressed as follows:
\begin{equation}
\begin{matrix}X_{i+1}^{\left(1\right)}=W_i^{\left(1\right)}\cdot X_i^{\left(0\right)}+b_i^{\left(1\right)}\\{\hat{X}}_{i+1}^{\left(1\right)}={\hat{W}}_i^{\left(1\right)}\cdot X_i^{\left(0\right)}+{\hat{b}}_i^{\left(1\right)}\\\begin{matrix}X_i^{\left(2\right)}=H_i\cdot\left(\sigma\left(W_i^{\left(2\right)}\cdot X_i^{\left(1\right)}\right)+b_i^{\left(2\right)}\right)+{\hat{X}}_{i+1}^{\left(1\right)}\\F_i=\mathcal{F}_i\odot X_i^{\left(2\right)}+X_{i-1}^{\left(2\right)}\\\end{matrix}\\\end{matrix}
\end{equation}
Here, $X_i^{(0)}$ denotes the input feature map to the $i$-th stage encoder. $W_i^{(1)},W_i^{(2)},b_i^{(1)},b_i^{(2)}$ represent the learnable weight matrices in the primary encoder, while ${\hat{W}}_i^{(1)},\ {\hat{b}}_i^{(1)}$ correspond to the learnable weight matrices in the auxiliary encoder. $\sigma$ signifies the activation function, $H_i$ denotes the cross-window self-attention function, $\mathcal{F}_i$ represents the intermediate transformation function, and $\odot$ indicates element-wise multiplication.
\begin{figure*}
    \centering
    \includegraphics[width=1\linewidth]{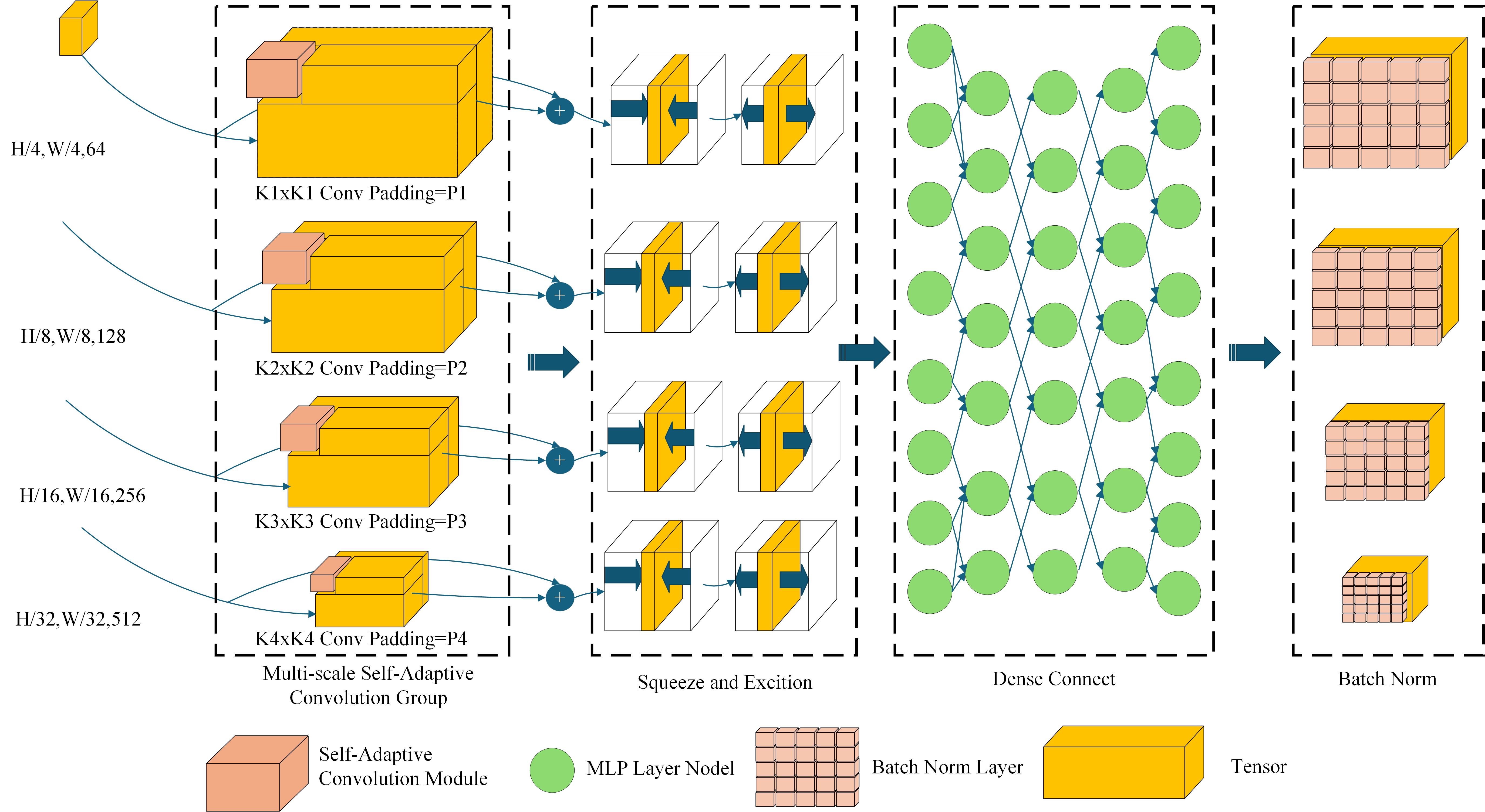}
    \caption{MSConvBrdige Architecture Diagram}
    \label{fig:4}
\end{figure*}
\subsection{MSConvBridge}
Although the encoder can effectively capture multi-scale information, directly feeding the original feature maps to the decoder may introduce redundant noise, thereby compromising segmentation accuracy. Convolutional operations function analogously to filters, capable of suppressing redundant information in feature maps while preserving critical features when detecting sub-structures. To enhance semantic consistency between the encoder and decoder, MSA\textsuperscript{2}-Net incorporates a Multi-Scale Convolutional Bridge (MSConvBridge), which selectively refines and optimizes features within the skip connections. The operational mechanism of the MSConvBridge is detailed below.
For a given feature map $F_i\in\mathbb{R}^{H\times W\times C}$ output from the $i$-th stage encoder, it first undergoes processing via an adaptive convolution operation:

\begin{equation}
{\bar{F}}_i^{\left(conv\right)} = \text{Conv}_{k_g}\left(F_i\right), k_g \in {\hat{W}}_C
\end{equation}

\begin{figure}
    \centering
    \includegraphics[width=1\linewidth]{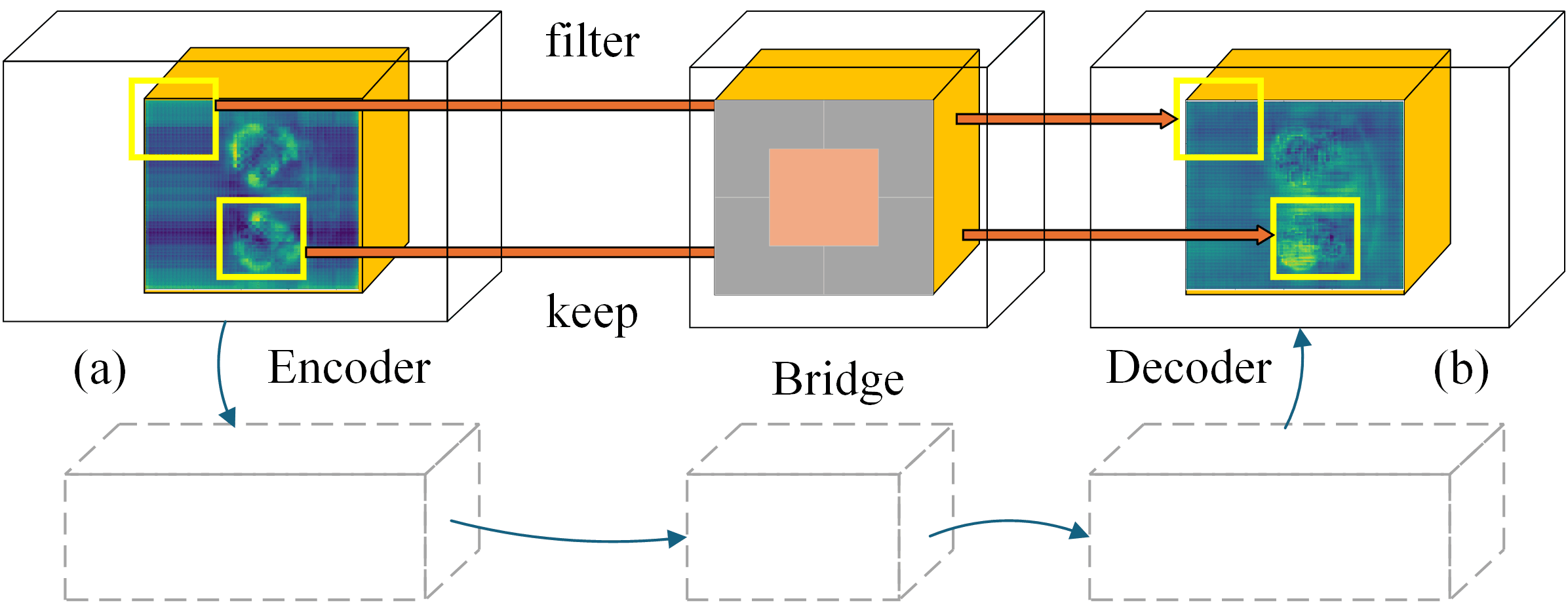}
    \caption{(a) CSWin feature maps without MSConvBridge processing. (b) CSWin feature maps after MSConvBridge processing.}
    \label{fig:5}
\end{figure}

The feature map ${\bar{F}}_i^{\left(g,conv\right)}$ undergoes the following operations in MSConvBridge:
\begin{equation}
\widehat{F}_{i}=H_{D C} \cdot H_{S E} \cdot\left(W_{i}^{o} \cdot \bar{F}_{i}^{(\text {conv })}+b_{i}^{o}\right), i \in[1,4]
\end{equation}
Here, $W_i^o, b_i^o$ represent the learnable weight matrices of the Self-Adaptive Convolution Module within the MSConvBridge, $H_{SE}$ denotes the squeeze-and-excitation function, and $H_{DC}$ indicates the dense connectivity function. As shown in Figure \ref{fig:5}(a), the feature maps without MSConvBridge processing exhibit prominent banding artifacts. In contrast, Figure \ref{fig:5}(b) demonstrates the output feature maps after processing by the MSConvBridge, where redundant noise is visibly suppressed.
\subsection{MSADecoder}
To restore spatial resolution and accurately reconstruct fine structures such as organ boundaries and small organs, MSA\textsuperscript{2}-Net employs a lightweight yet highly expressive decoding module—MSADecoder. As shown in Figure 6, the feature map $\widehat{F_i}\in\mathbb{R}^{H\times W\times C}$ is split into G parts along the channel dimension, yielding $\left\{{\hat{F}}_i^{\left(g\right)}\right\}_{g=1}^G$, where ${\hat{F}}_i^{\left(g\right)}\in\mathbb{R}^{H\times W\times\frac{C}{G}}$ with $G$ set to 4. In the MSADecoder, $\left\{{\hat{F}}_i^{\left(g\right)}\right\}_{g=1}^G$ first undergoes parallel processing through $G$ adaptive convolution operations:
\begin{equation}
{\hat{F}}_i^{\left(g,conv\right)}={Conv}_{k_g}\left({\hat{F}}_i^{\left(g\right)}\right),k_g\in{\hat{W}}_C
\end{equation}
${\hat{F}}_i^{\left(g,conv\right)}$ is then enhanced through a squeeze-and-excitation network to strengthen the internal feature representations:
\begin{equation}
y_i=H_E\cdot\left(w_i^\xi\cdot\left(H_S\cdot\left[{\hat{F}}_i^{\left(g,conv\right)}\right]_{g=1}^G\right)+b_i^\xi\right)
\end{equation}
Here, $H_S$ and $H_E$ denote the transformation matrices of the squeeze-and-excitation network in the MSADecoder, while $w_i^\xi$ and $b_i^\xi$ represent the learnable weight matrices of the Self-Adaptive Convolution Module in MSADeocder. The multi-scale fusion decoding strategy enables the MSADeocder to progressively restore spatial resolution while preserving fine details of small targets and boundary integrity, thereby significantly improving final segmentation accuracy.
\begin{figure*}
    \centering
    \includegraphics[width=1\linewidth]{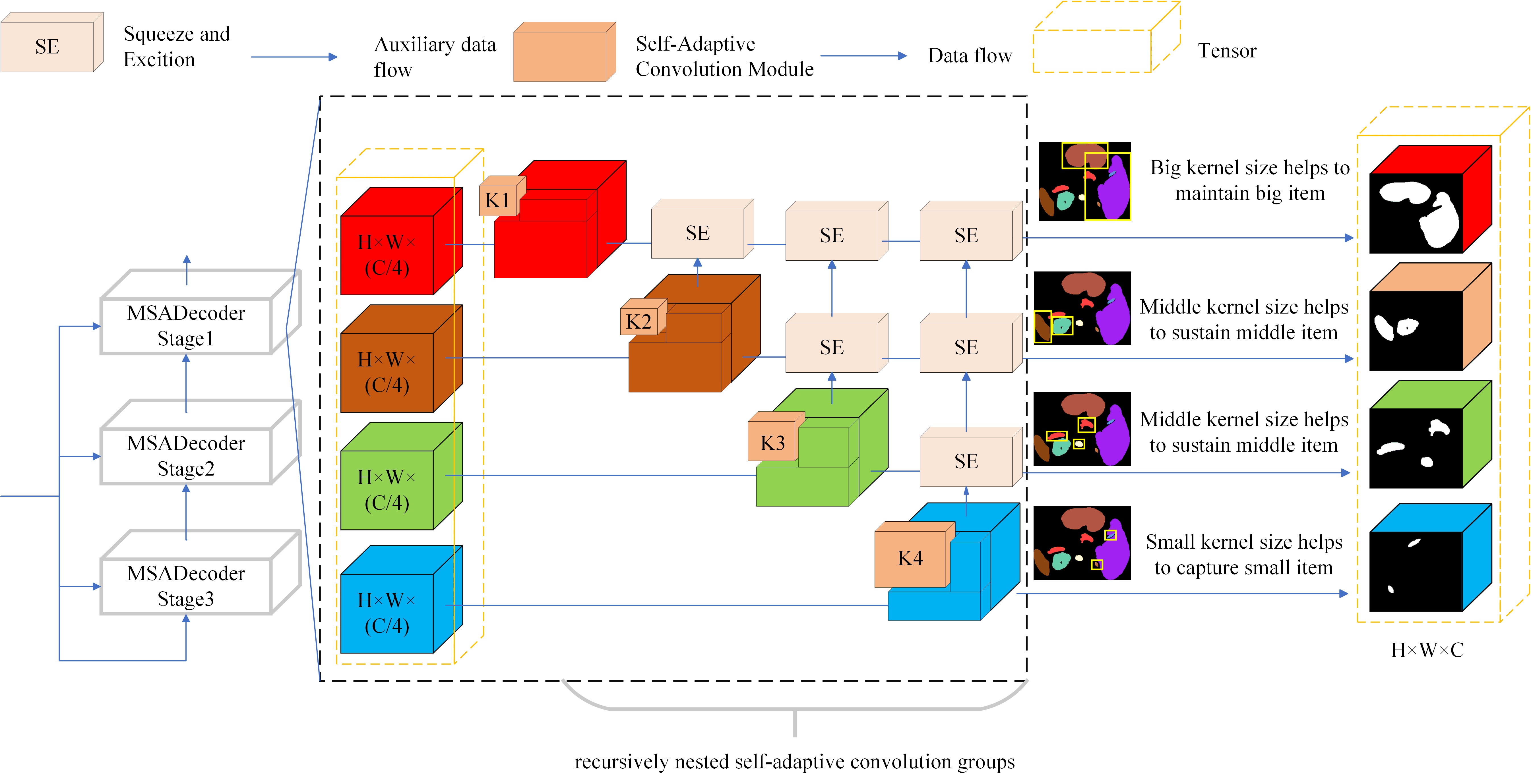}
    \caption{MSADecoder Architecture Diagram}
    \label{fig:6}
\end{figure*}
\section{Experiments}
\subsection{Environmental}
MSA\textsuperscript{2}-Net is implemented using PyTorch and trained on an NVIDIA A100 GPU platform. The CSWin backbone network incorporates pre-trained weights. Input images for the CSWin encoder are resized to $224\times244$ pixels. The initial learning rate is set to 0.0001 with a maximum of 300 epochs and a batch size of 14. The AdamW optimizer's weight decay is configured at 0.0001.
\subsection{Dataset}
In this section, we benchmark MSA\textsuperscript{2}-Net against current state-of-the-art (SOTA) networks to demonstrate its superiority. Experiments are conducted on three datasets: the Synapse multi-organ segmentation dataset (Synapse), the Automated Cardiac Diagnosis Challenge dataset (ACDC), and Kvasir. Network performance is evaluated using the Dice coefficient and average Hausdorff Distance (HD95). Results for comparative networks are obtained from previously published studies.
\subsubsection{Synapse}
Synapse dataset is a multi-organ segmentation benchmark comprising 30 abdominal CT scans with 3779 axial clinical CT images. Each CT volume consists of 85–198 slices of $512\times512$ pixels. It contains annotations for eight abdominal organs (aorta, gallbladder, spleen, left kidney, right kidney, liver, pancreas, and stomach), randomly divided into 18 scans for training and 12 for testing. To ensure training consistency, we adopt the Synapse dataset preprocessed by TransUNet\cite{RN47}.
\subsubsection{ACDC}
The ACDC dataset comprises 100 MRI scans collected from different patients, with each scan annotated for three organs: the left ventricle (LV), right ventricle (RV), and myocardium (MYO). The ACDC dataset contains cardiac examination results from different patients acquired via MRI scanners. The MR images were captured during breath-holding, with a series of short-axis slices covering the heart from the left ventricular base to the apex, featuring a slice thickness of 5 to 8 mm. The in-plane spatial resolution of the short-axis images ranges from $\left(0.83-1.75\right)mm^2/pixel$. For each patient scan, the ground truth of the left ventricle (LV), right ventricle (RV), and myocardium (MYO) was manually annotated. The ACDC dataset is randomly divided into 70 training cases (1,930 axial slices)\cite{RN48}, 10 cases for validation, and 20 cases for testing. Consistent with previous works, this study employs the Dice coefficient to evaluate the performance of MSA\textsuperscript{2}-Net.
\subsubsection{Kvasir-SEG}
Kvasir-SEG is an endoscopic dataset designed for pixel-level segmentation of colorectal polyps. It consists of 1,000 gastrointestinal polyp images along with their corresponding segmentation masks, all meticulously annotated and verified by experienced gastroenterologists. The dataset provides a standardized split of 880 images for training and 120 for validation to ensure fair comparison among different methods.
This dataset aims to advance research and innovation in the segmentation, detection, localization, and classification of colorectal polyps. Following the same approach as previous studies\cite{RN49}, we use 70 cases for training, 10 for validation, and 20 for testing.
\subsubsection{Skin Lesion Segmentation}
This study employs three public datasets—ISIC2017, ISIC2018, and PH²—for skin lesion segmentation experiments. The ISIC datasets contain a large number of dermoscopic images covering various types of lesions, with the specific data splits as follows: ISIC2017: Divided into 1,400 training images, 200 validation images, and 400 test images. ISIC2018: Consists of 1,815 training images, 259 validation images, and 520 test images. PH²: Configured with 80 training images, 20 validation images, and 100 test images\cite{RN61}.
The segmentation performance is evaluated using four metrics: mean Dice coefficient (Dice), sensitivity (SE), specificity (SP), and accuracy (ACC).
\subsection{Evaluation Metrics}
To comprehensively evaluate the segmentation performance of the proposed network, this study employs five key metrics:
\subsubsection{Dice}
Dice is the most widely used evaluation metric in medical imaging competitions. As a set similarity measure, it is typically employed to calculate the similarity between two samples, with a value range of $[0,1]$. In medical imaging, it is commonly used for image segmentation, where a score of 1 represents the best possible result and 0 indicates the worst. The formula for the Dice coefficient is as follows:
\begin{equation}
    Dice=\frac{2\times\left(pred\cap t r u e\right)}{pred\cup t r u e}
\end{equation}
In the formula, $pred$ indicates the point set generated by the network's prediction, while $true$ corresponds to the ground truth point set.
\subsubsection{HD95}
The 95\% Hausdorff distance (HD95) is used in this experiment to measure the degree of boundary overlap between two sets. It calculates the distance between the boundaries of the predicted segmentation and the ground truth, with smaller values indicating greater similarity between the two sets. The formula for HD95 is derived from the Hausdorff distance and is defined as follows:

\begin{equation}
    \begin{aligned}\label{eq:eq9}
        & \mathcal{D}_{HD95}=\left\{X,Y\right\}=max\left\{\mathcal{D}_{XY},\mathcal{D}_{YX}\right\}\\
        & \mathcal{D}_{XY}=\max_{x \in X}\left(\min_{y \in Y}\left(d\left(x,y\right)\right)\right)\\
        & \mathcal{D}_{YX}=\max_{y \in Y}\left(\min_{x \in X}\left(d\left(x,y\right)\right)\right)
    \end{aligned}
\end{equation}

Where $X$ and $Y$ represent two sets, $\mathcal{D}_{XY}$ represents the maximum distance from the point in set $X$ to the nearest point in set $Y$, $\mathcal{D}_{YX}$ represents the maximum distance from the point in set $Y$ to the nearest point in set $X$, and $\mathcal{D}_{HD95}$ represents the 95\% Hausdorff distance between the two sets.\par
\subsubsection{Sensitivity}
Sensitivity is a statistical metric that measures the ability of a model or test to correctly identify positive cases. It quantifies the proportion of actual positives that are correctly detected by the system. The calculation formula for Sensitivity is as follows:
\begin{equation}
    Sensitivity=\frac{TP}{TP+FN}
\end{equation}
\subsubsection{Specificity}
Specificity is a statistical metric that measures a model's ability to correctly identify negative cases. It quantifies the proportion of actual negatives that are correctly classified as such. The calculation formula for Specificity is as follows:
\begin{equation}
    Specificity=\frac{TN}{TN+FP}
\end{equation}
\subsubsection{Accuracy}
Accuracy is a fundamental performance metric that measures the overall correctness of a classification model. It represents the ratio of correctly predicted observations (both true positives and true negatives) to the total number of observations. The calculation formula for Accuracy is as follows:
\begin{equation}
    Accuracy=\frac{TP+TN}{TP+TN+FP+FN}
\end{equation}

\subsection{Experimental results and analysis}

In this section, we will compare the performance of MSA\textsuperscript{2}-Net with other state-of-the-art (SOTA) methods on the Synapse, ACDC, Skin Lesion Segmentation Datasets and Kvasir-SEG datasets. The experimental results are presented in Table \ref{tab:1}, Table \ref{tab:2}, and Table \ref{tab:3}.\par
\subsubsection{Performance of different networks on Synapse}
\begin{table*}[htbp]
  \centering
  \caption{Performance of different networkss on Synapse}
    \begin{tabular}{ccccccccccc}
    \toprule
    \multirow{2}[2]{*}{Architectures} & \multicolumn{2}{c}{Average} & \multirow{2}[2]{*}{Aorta} & \multirow{2}[2]{*}{GB} & \multirow{2}[2]{*}{KL} & \multirow{2}[2]{*}{KR} & \multirow{2}[2]{*}{Liver} & \multirow{2}[2]{*}{PC} & \multirow{2}[2]{*}{SP} & \multirow{2}[2]{*}{SM} \\
          & Dice↑ & HD95a↓ &       &       &       &       &       &       &       &  \\
    \midrule
    UNet\cite{RN4} (2015) & 70.11 & 44.69 & 84.00    & 56.7  & 72.41 & 62.64 & 86.98 & 48.73 & 81.48 & 67.96 \\
    AttnUNet\cite{RN60} (2018) & 71.70  & 34.47 & 82.61 & 61.94 & 76.07 & 70.42 & 87.54 & 46.70  & 80.67 & 67.66 \\
    R50+UNet\cite{RN7} (2021) & 74.68 & 36.87 & 84.18 & 62.84 & 79.19 & 71.29 & 93.35 & 48.23 & 84.41 & 73.92 \\
    R50+AttnUNet\cite{RN7} (2021) & 75.57 & 36.97 & 55.92 & 63.91 & 79.2  & 72.71 & 93.56 & 49.37 & 87.19 & 74.95 \\
    TransUNet\cite{RN7} (2021) & 77.48 & 31.69 & 87.23 & 63.13 & 81.87 & 77.02 & 94.08 & 55.86 & 85.08 & 75.62 \\
    SSFormerPVT\cite{RN52} (2022) & 78.01 & 25.72 & 82.78 & 63.74 & 80.72 & 78.11 & 93.53 & 61.53 & 87.07 & 76.61 \\
    PolypPVT\cite{RN53} ( 2021) & 78.08 & 25.61 & 82.34 & 66.14 & 81.21 & 73.78 & 94.37 & 59.34 & 88.05 & 79.40 \\
    MT-UNet\cite{RN54} (2022a) & 78.59 & 26.59 & 87.92 & 64.99 & 81.47 & 77.29 & 93.06 & 59.46 & 87.75 & 76.81 \\
    Swin-UNet\cite{RN32} (2021) & 79.13 & 21.55 & 85.47 & 66.53 & 83.28 & 79.61 & 94.29 & 56.58 & 90.66 & 76.60 \\
    PVT-CASCADE\cite{RN48} (2023) & 81.06 & 20.23 & 83.01 & 70.59 & 82.23 & 80.37 & 94.08 & 64.43 & 90.10  & 83.69 \\
    MISSFormer\cite{RN28} (2021) & 81.96 & 18.20  & 86.99 & 68.65 & 85.21 & 82.00    & 94.41 & 65.67 & 91.92 & 80.81 \\
    CASTformer\cite{RN55} (2022) & 82.55 & 22.73 & \textcolor[rgb]{ .933,  0,  0}{89.05} & 67.48 & 86.05 & 82.17 & 95.61 & 67.49 & 91.00    & 81.55 \\
    TransCASCADE\cite{RN48}(2023) & 82.68 & 17.34 & 86.63 & 68.48 & 87.66 & 84.56 & 94.43 & 65.33 & 90.79 & 83.52 \\
    Cascaded MERIT\cite{RN48} & 84.90  & \textcolor[rgb]{ .933,  0,  0}{13.22} & 87.71 & 74.40  & 87.79 & 84.85 & 95.26 & 71.81 & 92.01 & 85.38 \\
    AgileFormer\cite{RN55} & 85.74 & 18.70  & 89.11 & \textcolor[rgb]{ .933,  0,  0}{77.89} & \textcolor[rgb]{ .933,  0,  0}{88.83} & 85.00    & 95.64 & 71.62 & 92.2  & 85.63 \\
    MSA\textsuperscript{2}-Net(ours) & \textcolor[rgb]{ .933,  0,  0}{86.49} & 14.15 & 85.90  & 74.44 & 86.72 & \textcolor[rgb]{ .933,  0,  0}{86.77} & \textcolor[rgb]{ .933,  0,  0}{96.57} & \textcolor[rgb]{ .933,  0,  0}{82.34} & \textcolor[rgb]{ .933,  0,  0}{92.93} & \textcolor[rgb]{ .933,  0,  0}{86.31} \\
    \bottomrule
    \end{tabular}%
  \label{tab:1}%
\end{table*}%
Table \ref{tab:1} presents the performance of MSA\textsuperscript{2}-Net on the Synapse dataset, highlighting its superior capabilities. MSA\textsuperscript{2}-Net achieved the highest average Dice coefficient of 86.49\%, showing improvements of 5.52\%, 9.30\%, and 0.65\% over MISSFormer, Swin-UNet, and AgileFormer, respectively. In terms of the HD95 metric, MSA\textsuperscript{2}-Net recorded the second lowest distance (14.15), significantly outperforming MISSFormer (18.20), Swin-UNet (21.55), and TransCASCADE (17.34).\par
\begin{figure}
    \centering
    \includegraphics[width=1\linewidth]{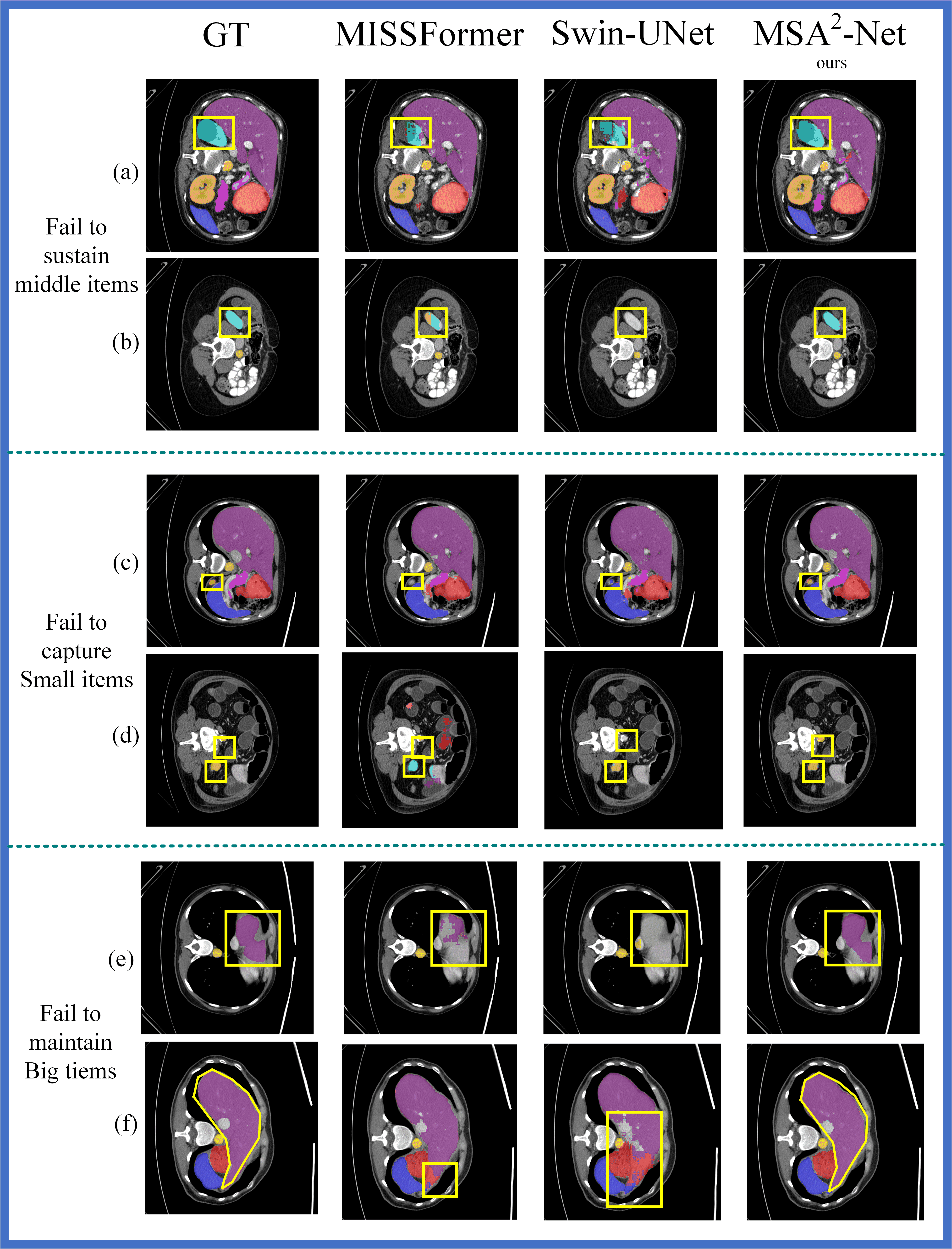}
    \caption{Segmentation effects of MSA\textsuperscript{2}-Net, MISSFormer, and SwinUnet on the Synapse}
    \label{fig:7}
\end{figure}

By examining Table \ref{tab:1}, it can be observed that MSA\textsuperscript{2}-Net achieved the best Dice scores for 5 out of 8 organs. These organs are all small in size, demonstrating that the MSA\textsuperscript{2}-Net efficiently preserves information for small organs during upsampling. Due to space constraints, and considering that MISSFormer is the primary inspiration for MSA\textsuperscript{2}-Net and SwinUnet is a classic algorithm in medical image segmentation, Figure \ref{fig:7} displays the segmentation results of MSA\textsuperscript{2}-Net, MISSFormer, and SwinUnet on the Synapse dataset.\par
Figure \ref{fig:7} provides a visualization of the segmentation task performed by MSA\textsuperscript{2}-Net, MISSFormer, and Swin-UNet on the Synapse dataset. In this figure, the first column displays the Ground Truth images, the second column shows the segmentation images generated by MISSFormer, the third column presents the images produced by Swin-UNet, and the fourth column depicts the segmentation images created by MSA\textsuperscript{2}-Net. The areas circled by the yellow line indicate the locations of sub-features where most methods failed to make accurate predictions.\par
Rows (a) and (b) in Figure \ref{fig:7} illustrate how other methods struggle to effectively capture medium-sized objects during the segmentation task on the Synapse dataset. In row (a), the area circled by the yellow box represents the left kidney. MSA\textsuperscript{2}-Net segments the left kidney most completely, while MISSFormer and Swin-UNet fail to segment it as effectively due to their inability to preserve multi-scale information. In row (b), the yellow box again highlights the left kidney, but in a different channel dimension, resulting in a smaller ground truth value for the left kidney. MISSFormer retains some features of the left kidney because of its fixed-size convolution operation. In contrast, Swin-UNet does not employ convolution operations to maintain sub-features during the up-sampling process, which prevents it from detecting the left kidney. MSA\textsuperscript{2}-Net successfully captures the multi-scale spatial information of the left kidney through the use of an MSADecoder, enabling it to correctly recognize the left kidney across different channel dimensions.\par
In rows (c) and (d), the area circled in yellow represents the gallbladder, a small organ. MISSFormer fails to capture the gallbladder because it uses a fixed-size convolution that filters out small features, while Swin-UNet misidentifies the gallbladder as other organs due to the lack of convolution operations during up-sampling. Only MSA\textsuperscript{2}-Net successfully captures the gallbladder.
In rows (e) and (f), the difficulties in maintaining large objects are observed. In row (f), the yellow box highlights the liver. Although MISSFormer partially restores the liver segmentation due to its fixed-size convolution, it cannot fully capture the liver because its convolution kernel is too small, limiting its receptive field. Swin-UNet shows a significant discrepancy between the reduced liver and the original labeling because it lacks multi-scale convolution capabilities. In contrast, MSA\textsuperscript{2}-Net, with its MSADecoder, completely restores the liver segmentation map, effectively handling the large object.\par
\subsubsection{Performance of different networks on the ACDC and Kvasir-SEG}
\begin{figure}
    \centering
    \includegraphics[width=1\linewidth]{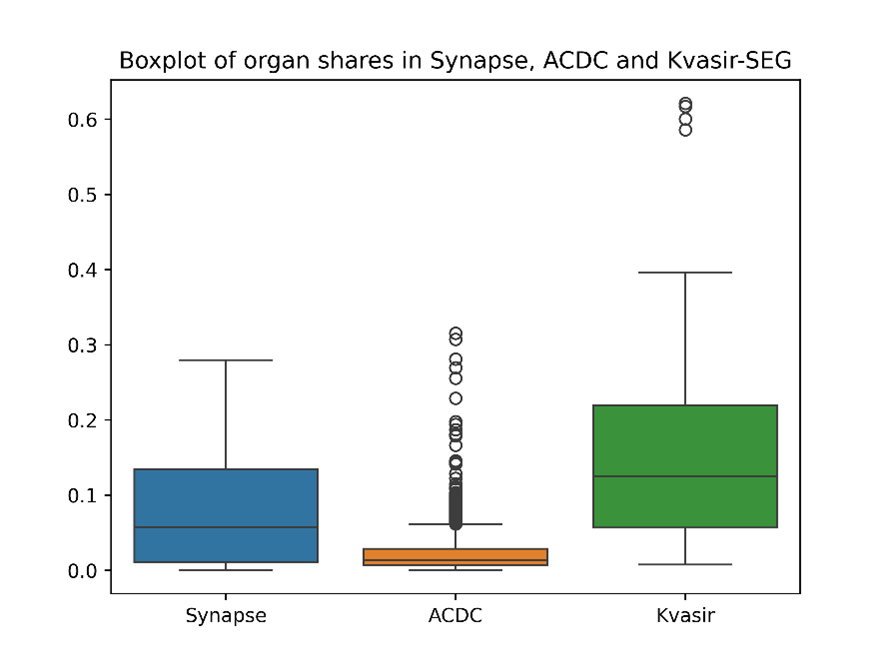}
    \caption{the box plots of the area proportions of organs in the Synapse, ACDC, and Kvasir-SEG datasets}
    \label{fig:8}
\end{figure}

\begin{table*}[htbp]
  \centering
  \caption{Analysis of the results of the ACDC and Kvasir-SEG}
    \begin{tabular}{ccccc|ccccc}
    \toprule
    \multicolumn{5}{c}{ACDC}              & \multicolumn{5}{c}{Kvasir-SEG} \\
    \midrule
    Architecture & Dice  & RV    & MYO   & LV    & Architecture & Dice  & SE    & SP    & ACC \\
    U-Net\cite{RN4} & 87.55 & 87.10  & 80.63 & 94.92 & BDG-Net\cite{RN63} & \textcolor[rgb]{ 1,  0,  0}{91.50} & \textcolor[rgb]{ 1,  0,  0}{68.53} & 92.74 & 74.11 \\
    Att-Unet\cite{RN60} & 86.75 & 87.58 & 79.20  & 93.47 & KDAS3\cite{solar2009identifying} & 91.30  & 66.03 & 93.93 & 76.25 \\
    nnUNet\cite{RN33} & 90.91 & 89.21 & \textcolor[rgb]{ 1,  0,  0}{90.20} & 93.35 & U-Net++\cite{wisaeng2023u} & 82.10  & 44.62 & 84.48 & 72.01 \\
    UNetR\cite{hatamizadeh2022unetr} & 88.61 & 85.29 & 86.52 & 94.02 & Polyp-SAM++\cite{biswas2023polyp} & 90.20  & 67.29 & \textcolor[rgb]{ 1,  0,  0}{96.82} & 73.56 \\
    TransUNet\cite{RN7} & 89.71 & 88.86 & 84.53 & 95.73 & U-Net\cite{RN4} & 81.80  & 64.28 & 86.90  & 70.21 \\
    Swin-UNet\cite{RN32} & 90.00    & 88.55 & 85.63 & \textcolor[rgb]{ 1,  0,  0}{95.83} & TGA-Net\cite{liu2022connecting} & 89.82 & 62.53 & 84.89 & 75.43 \\
    CSwin-UNet\cite{RN20} & 91.46 & 89.68 & 88.94 & 95.76 & PEFNet\cite{RN66} & 88.18 & 59.98 & 76.32 & 66.34 \\
    ST-UNet\cite{yu2019st} & 89.73 & 87.65 & 82.11 & 94.39 & TransNetR\cite{RN67} & 87.06 & 56.45 & 79.54 & 78.67 \\
    UNetFormer\cite{wang2022unetformer} & 89.09 & 88.92 & 87.88 & 95.03 & ResUNet++\cite{RN68} & 81.33 & 44.60  & 77.30  & 72.66 \\
    MSA\textsuperscript{2}-Net(ours) & \textcolor[rgb]{ 1,  0,  0}{92.56} & \textcolor[rgb]{ 1,  0,  0}{90.95} & 89.09 & 95.63 & MSA\textsuperscript{2}-Net(ours) & 91.49 & 62.14 & 94.24 & \textcolor[rgb]{ 1,  0,  0}{78.88} \\
    \bottomrule
    \end{tabular}%
  \label{tab:2}%
\end{table*}%
Table \ref{tab:2} presents the experimental results of MSA\textsuperscript{2}-Net on the ACDC and Kvasir-SEG datasets. In the ACDC dataset, MSA\textsuperscript{2}-Net achieved the best results compared to other SOTA networks, with Dice score improvements of 1.08\% and 1.88\% over MISSFormer and Swin-UNet, respectively. Notably, MSA\textsuperscript{2}-Net achieved the highest score of 90.95\% in the RV subtask. This indicates that MSA\textsuperscript{2}-Net excels across various medical imaging data modalities. However, MSA\textsuperscript{2}-Net's performance on the Kvasir-SEG dataset was less impressive. To explore this, we examined the box plots in Figure \ref{fig:8}, which show the area proportions of organs in the Synapse, ACDC, and Kvasir-SEG datasets.\par

Figure \ref{fig:8} shows that the organ area proportions in the Kvasir-SEG dataset have a wider distribution compared to those in the Synapse and ACDC datasets. (Note: Each box plot in Figure \ref{fig:8} contains outliers, which are actual organ area proportions and were not removed during plotting.) Since the Kvasir-SEG dataset only contains "tumor" as the target organ for segmentation, it indicates that the shape and area of "tumors" vary significantly. This demonstrates that the current MSA\textsuperscript{2}-Net still lacks the capability to handle organs with highly variable shapes.\par
\subsubsection{Experimental Results on Skin Lesion Segmentation Datasets}

\begin{table*}[htbp]
  \centering
  \caption{Experimental Results of MSA\textsuperscript{2}-Net on Skin Lesion Segmentation Datasets}
    \begin{tabular}{ccccc|cccc|cccc}
    \toprule
    \multirow{2}[4]{*}{Method} & \multicolumn{4}{c}{ISIC2017}  & \multicolumn{4}{c}{ISIC2018}  & \multicolumn{4}{c}{PH2} \\
\cmidrule{2-13}          & Dice  & SE    & SP    & \multicolumn{1}{c}{ACC} & Dice  & SE    & SP    & \multicolumn{1}{c}{ACC} & Dice  & SE    & SP    & ACC \\
    \midrule
    U-Net\cite{RN4} & 81.59 & 81.72 & 96.80  & 91.64 & 85.45 & 88.00    & 96.97 & 94.04 & 89.36 & 91.25 & 95.88 & 92.33 \\
    Att-Unet\cite{RN60} & 80.82 & 79.98 & 97.76 & 91.45 & 85.66 & 86.74 & 98.63 & 93.76 & 90.03 & 92.05 & 96.40  & 92.76 \\
    TransUNet\cite{RN7} & 81.23 & 82.63 & 95.77 & 92.07 & 84.99 & 85.78 & 96.53 & 94.52 & 88.40  & 90.63 & 94.27 & 92.00 \\
    HiFormer\cite{RN61} & 92.53 & 91.55 & 98.4  & 97.02 & 91.02 & 91.19 & 97.55 & 96.21 & 94.60  & 94.20  & 97.72 & 96.61 \\
    Swin-UNet\cite{RN32} & 91.83 & 91.42 & 97.98 & 97.01 & 89.46 & 90.56 & 97.98 & \textcolor[rgb]{ 1,  0,  0}{96.45} & 94.49 & 94.10  & 95.64 & 96.78 \\
    MISSFormer\cite{RN28} & 89.03 & 89.24 & 97.25 & 95.69 & 91.01 & 90.31 & 97.45 & 94.42 & 94.01 & 93.05 & 96.91 & 96.14 \\
    TMU\cite{reza2022contextual}   & 91.64 & 91.28 & 97.89 & 96.60  & 90.59 & 90.38 & 97.46 & 96.03 & 91.14 & 93.95 & 97.56 & 96.47 \\
    CSwin-UNet\cite{RN20} & 91.47 & 93.79 & \textcolor[rgb]{ 1,  0,  0}{98.56} & \textcolor[rgb]{ 1,  0,  0}{97.26} & 91.11 & \textcolor[rgb]{ 1,  0,  0}{92.31} & 97.88 & 95.25 & 94.29 & \textcolor[rgb]{ 1,  0,  0}{95.63} & 97.82 & 96.82 \\
    
    MSA\textsuperscript{2}-Net(ours) & \textcolor[rgb]{ 1,  0,  0}{92.98} & \textcolor[rgb]{ 1,  0,  0}{93.85} & 97.25 & 97.01 & \textcolor[rgb]{ 1,  0,  0}{91.32} & 91.25 & \textcolor[rgb]{ 1,  0,  0}{98.36} & 96.12 & \textcolor[rgb]{ 1,  0,  0}{94.65} & \textcolor[rgb]{ 1,  0,  0}{95.65} & \textcolor[rgb]{ 1,  0,  0}{97.86} & \textcolor[rgb]{ 1,  0,  0}{96.85} \\
    \bottomrule
    \end{tabular}%
  \label{tab:3}%
\end{table*}%

\begin{figure}
    \centering
    \includegraphics[width=1\linewidth]{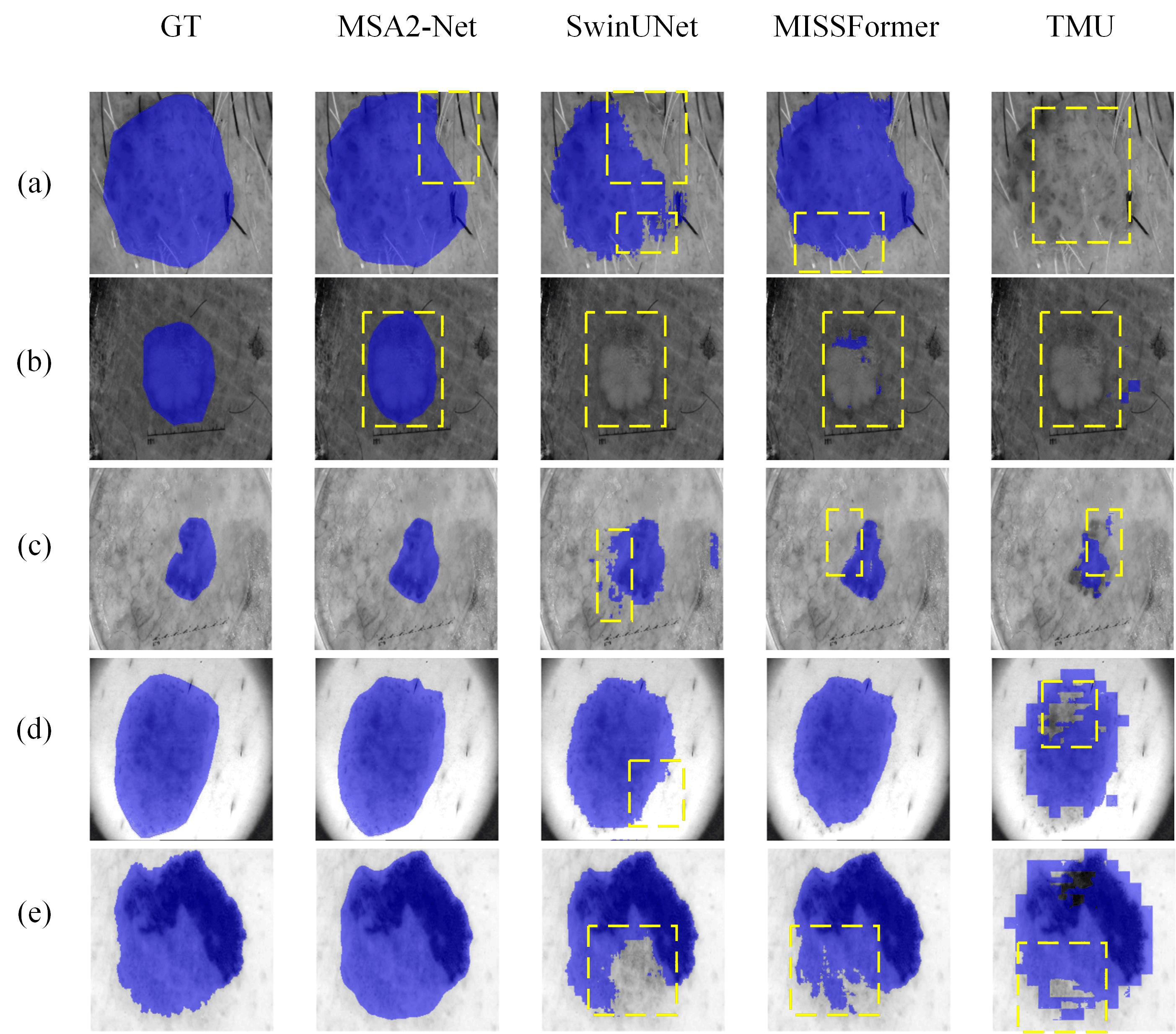}
    \caption{Visualization of MSA\textsuperscript{2}-Net Segmentation Results on the ISIC2017 Dataset}
    \label{fig:9}
\end{figure}

Table \ref{tab:3} shows the experimental results of MSA\textsuperscript{2}-Net on the ISIC dataset. In all versions of the ISIC dataset, MSA\textsuperscript{2}-Net achieved the best Dice scores, reflecting its strong segmentation performance and ability to accurately delineate skin lesion areas. Figure \ref{fig:9} presents the visual comparison of lesion segmentation by MSA\textsuperscript{2}-Net and other advanced models on the ISIC2017 dataset. Rows (a), (b), and (c) contain lesions with complex backgrounds, testing the network's ability to filter redundant information. Rows (d) and (e) have lesions with simple backgrounds, testing the network's ability to capture detailed information. In row (a), MSA\textsuperscript{2}-Net successfully identified most of the lesion area without being misled by background hair. In row (d), although all networks identified the main lesion area, MSA\textsuperscript{2}-Net captured the most detailed lesion boundaries.\par
\subsection{Ablation Study}
We conducted ablation experiments on the Synapse dataset to evaluate MSA\textsuperscript{2}-Net. Specifically, we explored the performance improvements brought by the adaptive convolution module, the MSConvBridge, and the MSADecoder.\par
\subsubsection{MSADecoder and MSConvBridge}

\begin{table}[htbp]
  \centering
  \caption{Ablation Study Results}

    \begin{tabular}{cccc}
    \toprule
     MSADecoder & MSConvBridge & $Dice\uparrow$ & $HD95_{a}\downarrow$ \\
    \midrule
     $\checkmark$     & $\checkmark$     & 86.49 & 14.15 \\
 $\checkmark$     & $\times$      & 78.56 & 22.43 \\
   $\times$   & $\checkmark$     & 77.90  & 19.65 \\
   $ \times$   &   $\times$    & 77.75 & 21.64 \\
\midrule
    \end{tabular}%

  \label{tab:4}%
\end{table}%
In this section, we performed ablation experiments on MSA\textsuperscript{2}-Net using the Synapse dataset, as shown in Table \ref{tab:4}. The results indicate that MSA\textsuperscript{2}-Net achieves optimal performance with both the multi-scale amalgamation decoder and the MSConvBridge, obtaining a Dice score of 86.49\% and an HD95 distance of 14.13 on the Synapse dataset. When MSA\textsuperscript{2}-Net is equipped with only the MSADecoder, the Dice score decreases by 9.16\% and HD95 increases by 8.28. When equipped with only the MSConvBirdge, the Dice score decreases by 9.93\% and HD95 increases by 5.5. Without both the MSConvBirdge and the MSADecoder, the Dice score decreases by 10.10\% and HD95 increases by 7.49.\par
\subsection{Self-Adaptive convolution Module}

\begin{table}[htbp]
  \centering
  \caption{Impact of Different Guidance Methods on Experimental Results}
  \small

    \resizebox{0.45\textwidth}{!}{ 
    \begin{tabular}{cc|ccccc}
    \toprule
    \multicolumn{2}{c}{Guidance}& Q1    & Q2    & Q3    & None  & Self-Adaptive \\
    \midrule
     \multicolumn{2}{c|}{MSConvBridge} & [1,1,1,3] & [1,1,3,5] & [1,3,3,7] & [1,3,5,7] & [1,3,5,7] \\
 \multirow{3}[2]{*}{MSADecoder} & Stage1 & [1,1,1,3] & [1,1,3,5] & [1,3,3,7] & [1,3,5,7] & [1,3,3,7] \\
                 & Stage2 & [1,1,1,1] & [1,1,3,3] & [1,1,3,3] & [1,3,5,7] & [1,1,3,3] \\
                 & Stage3 & [1,1,1,1] & [1,1,1,1] & [1,1,3,3] & [1,3,5,7] & [1,1,3,3] \\
 \multicolumn{2}{c|}{Synapse} & 82.03 & 82.74 & 86.12 & 84.53 & \textcolor[rgb]{ 1,  0,  0}{86.49} \\
 \multicolumn{2}{c|}{ACDC} & 90.11 & 90.86 & 91.88 & 91.07 & \textcolor[rgb]{ 1,  0,  0}{92.54} \\
 \multicolumn{2}{c|}{ISIC2017} & 77.64 & 89.98 & 92.01 & 91.78 & \textcolor[rgb]{ 1,  0,  0}{92.34} \\
 \multicolumn{2}{c|}{ISIC2018} & 78.79 & 88.22 & 91.21 & 90.85 & \textcolor[rgb]{ 1,  0,  0}{91.32} \\
 \multicolumn{2}{c|}{Kvasir-SEG} & 64.51 & 82.19 & \textcolor[rgb]{ 1,  0,  0}{91.75} 
& 89.09 & 91.49 \\
\midrule
    \end{tabular}%
    }

  \label{tab:tab5}%
\end{table}%
Table \ref{tab:tab5} shows the impact of different guidance principles on the self-adaptive convolution module. In these principles, Q1, Q2, and Q3 use quartile-based weighting for all adaptive convolution modules, None uses the original convolution kernel scheme of the IDConv module, and Self-Adaptive refers to the self-adaptive convolution module proposed in this study. It can be seen that the self-adaptive module achieves the best results in the Synapse, ACDC, and ISIC datasets.\par
\section{Conclusion}
This work introduces the design concept and application of the self-adaptive convolution module, as well as the application of the MSADecoder and MSConvbridge in medical image segmentation. The self-adaptive convolution module can adjust the size of the convolution kernels based on the characteristics of the dataset. MSConvbridge filters the decoder's redundant outputs while modeling global and local information within feature maps. MSADecoder fully utilizes multi-scale information during upsampling to fill in missing details within feature maps. MSA\textsuperscript{2}-Net has demonstrated its superior performance on four prominent medical datasets: Synapse, ACDC, ISIC, and Kvasir-SEG. However, the experimental results on the Kvasir-SEG dataset indicate that MSA\textsuperscript{2}-Net still has limitations in handling organs with diverse morphologies, which is a direction for future improvement.\par
\bibliographystyle{IEEEtran} 
\bibliography{mintref} 

\begin{thebibliography}{10}
\providecommand{\url}[1]{#1}
\csname url@samestyle\endcsname
\providecommand{\newblock}{\relax}
\providecommand{\bibinfo}[2]{#2}
\providecommand{\BIBentrySTDinterwordspacing}{\spaceskip=0pt\relax}
\providecommand{\BIBentryALTinterwordstretchfactor}{4}
\providecommand{\BIBentryALTinterwordspacing}{\spaceskip=\fontdimen2\font plus
\BIBentryALTinterwordstretchfactor\fontdimen3\font minus \fontdimen4\font\relax}
\providecommand{\BIBforeignlanguage}[2]{{%
\expandafter\ifx\csname l@#1\endcsname\relax
\typeout{** WARNING: IEEEtran.bst: No hyphenation pattern has been}%
\typeout{** loaded for the language `#1'. Using the pattern for}%
\typeout{** the default language instead.}%
\else
\language=\csname l@#1\endcsname
\fi
#2}}
\providecommand{\BIBdecl}{\relax}
\BIBdecl

\bibitem{RN1}
S.~Asgari~Taghanaki, K.~Abhishek, J.~P. Cohen, J.~Cohen-Adad, and G.~Hamarneh, ``Deep semantic segmentation of natural and medical images: a review,'' \emph{Artificial Intelligence Review}, vol.~54, pp. 137--178, 2021.

\bibitem{RN2}
I.~Qureshi, J.~Yan, Q.~Abbas, K.~Shaheed, A.~B. Riaz, A.~Wahid, M.~W.~J. Khan, and P.~Szczuko, ``Medical image segmentation using deep semantic-based methods: A review of techniques, applications and emerging trends,'' \emph{Information Fusion}, vol.~90, pp. 316--352, 2023.

\bibitem{RN3}
R.~Wang, T.~Lei, R.~Cui, B.~Zhang, H.~Meng, and A.~K. Nandi, ``Medical image segmentation using deep learning: A survey,'' \emph{IET image processing}, vol.~16, no.~5, pp. 1243--1267, 2022.

\bibitem{RN5}
J.~Long, E.~Shelhamer, and T.~Darrell, ``Fully convolutional networks for semantic segmentation,'' in \emph{Proceedings of the IEEE conference on computer vision and pattern recognition}, Conference Proceedings, pp. 3431--3440.

\bibitem{RN6}
R.~Azad, A.~R. Fayjie, C.~Kauffmann, I.~Ben~Ayed, M.~Pedersoli, and J.~Dolz, ``On the texture bias for few-shot cnn segmentation,'' in \emph{Proceedings of the IEEE/CVF winter conference on applications of computer vision}, Conference Proceedings, pp. 2674--2683.

\bibitem{RN7}
J.~Chen, Y.~Lu, Q.~Yu, X.~Luo, E.~Adeli, Y.~Wang, L.~Lu, A.~L. Yuille, and Y.~Zhou, ``Transunet: Transformers make strong encoders for medical image segmentation,'' \emph{arXiv preprint arXiv:2102.04306}, 2021.

\bibitem{RN8}
M.~Wen, Q.~Zhou, B.~Tao, P.~Shcherbakov, Y.~Xu, and X.~Zhang, ``Short‐term and long‐term memory self‐attention network for segmentation of tumours in 3d medical images,'' \emph{CAAI Transactions on Intelligence Technology}, vol.~8, no.~4, pp. 1524--1537, 2023.

\bibitem{RN28}
X.~Huang, Z.~Deng, D.~Li, and X.~Yuan, ``Missformer: An effective medical image segmentation transformer,'' \emph{arXiv preprint arXiv:2109.07162}, 2021.

\bibitem{RN39}
G.~Xu, H.~Cao, J.~K. Udupa, Y.~Tong, and D.~A. Torigian, ``Disegnet: A deep dilated convolutional encoder-decoder architecture for lymph node segmentation on pet/ct images,'' \emph{Computerized Medical Imaging and Graphics}, vol.~88, p. 101851, 2021.

\bibitem{RN42}
L.~Chen, H.~Zhang, J.~Xiao, L.~Nie, J.~Shao, W.~Liu, and T.-S. Chua, ``Sca-cnn: Spatial and channel-wise attention in convolutional networks for image captioning,'' in \emph{Proceedings of the IEEE conference on computer vision and pattern recognition}, Conference Proceedings, pp. 5659--5667.

\bibitem{fu2021adaptive}
L.~Fu, W.-b. Gu, Y.-b. Ai, W.~Li, and D.~Wang, ``Adaptive spatial pixel-level feature fusion network for multispectral pedestrian detection,'' \emph{Infrared Physics \& Technology}, vol. 116, p. 103770, 2021.

\bibitem{RN33}
F.~Isensee, P.~F. Jaeger, S.~A. Kohl, J.~Petersen, and K.~H. Maier-Hein, ``nnu-net: a self-configuring method for deep learning-based biomedical image segmentation,'' \emph{Nature methods}, vol.~18, no.~2, pp. 203--211, 2021.

\bibitem{RN22}
X.~Dong, J.~Bao, D.~Chen, W.~Zhang, N.~Yu, L.~Yuan, D.~Chen, and B.~Guo, ``Cswin transformer: A general vision transformer backbone with cross-shaped windows,'' in \emph{Proceedings of the IEEE/CVF conference on computer vision and pattern recognition}, Conference Proceedings, pp. 12\,124--12\,134.

\bibitem{RN26}
Z.~Zhou, M.~M. Rahman~Siddiquee, N.~Tajbakhsh, and J.~Liang, ``Unet++: A nested u-net architecture for medical image segmentation,'' in \emph{Deep Learning in Medical Image Analysis and Multimodal Learning for Clinical Decision Support: 4th International Workshop, DLMIA 2018, and 8th International Workshop, ML-CDS 2018, Held in Conjunction with MICCAI 2018, Granada, Spain, September 20, 2018, Proceedings 4}.\hskip 1em plus 0.5em minus 0.4em\relax Springer, Conference Proceedings, pp. 3--11.

\bibitem{qi2018age}
Q.~Qi, B.~Du, M.~Zhuang, Y.~Huang, and X.~Ding, ``Age estimation from mr images via 3d convolutional neural network and densely connect,'' in \emph{International Conference on Neural Information Processing}.\hskip 1em plus 0.5em minus 0.4em\relax Springer, 2018, pp. 410--419.

\bibitem{RN47}
S.~Fu, Y.~Lu, Y.~Wang, Y.~Zhou, W.~Shen, E.~Fishman, and A.~Yuille, ``Domain adaptive relational reasoning for 3d multi-organ segmentation,'' in \emph{Medical Image Computing and Computer Assisted Intervention–MICCAI 2020: 23rd International Conference, Lima, Peru, October 4–8, 2020, Proceedings, Part I 23}.\hskip 1em plus 0.5em minus 0.4em\relax Springer, Conference Proceedings, pp. 656--666.

\bibitem{RN48}
M.~M. Rahman and R.~Marculescu, ``Medical image segmentation via cascaded attention decoding,'' in \emph{Proceedings of the IEEE/CVF Winter Conference on Applications of Computer Vision}, Conference Proceedings, pp. 6222--6231.

\bibitem{RN49}
N.~K. Tomar, A.~Shergill, B.~Rieders, U.~Bagci, and D.~Jha, ``Transresu-net: Transformer based resu-net for real-time colonoscopy polyp segmentation,'' \emph{arXiv preprint arXiv:2206.08985}, 2022.

\bibitem{RN61}
M.~Heidari, A.~Kazerouni, M.~Soltany, R.~Azad, E.~K. Aghdam, J.~Cohen-Adad, and D.~Merhof, ``Hiformer: Hierarchical multi-scale representations using transformers for medical image segmentation,'' in \emph{Proceedings of the IEEE/CVF winter conference on applications of computer vision}, 2023, pp. 6202--6212.

\bibitem{RN4}
O.~Ronneberger, P.~Fischer, and T.~Brox, ``U-net: Convolutional networks for biomedical image segmentation,'' in \emph{Medical image computing and computer-assisted intervention–MICCAI 2015: 18th international conference, Munich, Germany, October 5-9, 2015, proceedings, part III 18}.\hskip 1em plus 0.5em minus 0.4em\relax Springer, Conference Proceedings, pp. 234--241.

\bibitem{RN60}
X.-Z. Hu, W.-S. Jeon, and S.-Y. Rhee, ``Att-unet: Pixel-wise staircase attention for weed and crop detection,'' in \emph{2023 International Conference on Fuzzy Theory and Its Applications (iFUZZY)}.\hskip 1em plus 0.5em minus 0.4em\relax IEEE, 2023, pp. 1--5.

\bibitem{RN52}
J.~Wang, Q.~Huang, F.~Tang, J.~Meng, J.~Su, and S.~Song, ``Stepwise feature fusion: Local guides global,'' in \emph{International Conference on Medical Image Computing and Computer-Assisted Intervention}.\hskip 1em plus 0.5em minus 0.4em\relax Springer, Conference Proceedings, pp. 110--120.

\bibitem{RN53}
B.~Dong, W.~Wang, D.-P. Fan, J.~Li, H.~Fu, and L.~Shao, ``Polyp-pvt: Polyp segmentation with pyramid vision transformers,'' \emph{arXiv preprint arXiv:2108.06932}, 2021.

\bibitem{RN54}
H.~Wang, S.~Xie, L.~Lin, Y.~Iwamoto, X.-H. Han, Y.-W. Chen, and R.~Tong, ``Mixed transformer u-net for medical image segmentation,'' in \emph{ICASSP 2022-2022 IEEE international conference on acoustics, speech and signal processing (ICASSP)}.\hskip 1em plus 0.5em minus 0.4em\relax IEEE, Conference Proceedings, pp. 2390--2394.

\bibitem{RN32}
H.~Cao, Y.~Wang, J.~Chen, D.~Jiang, X.~Zhang, Q.~Tian, and M.~Wang, ``Swin-unet: Unet-like pure transformer for medical image segmentation,'' in \emph{European conference on computer vision}.\hskip 1em plus 0.5em minus 0.4em\relax Springer, Conference Proceedings, pp. 205--218.

\bibitem{RN55}
C.~You, R.~Zhao, F.~Liu, S.~Dong, S.~Chinchali, U.~Topcu, L.~Staib, and J.~Duncan, ``Class-aware adversarial transformers for medical image segmentation,'' \emph{Advances in neural information processing systems}, vol.~35, pp. 29\,582--29\,596, 2022.

\bibitem{RN63}
Z.~Qiu, Z.~Wang, M.~Zhang, Z.~Xu, J.~Fan, and L.~Xu, ``Bdg-net: boundary distribution guided network for accurate polyp segmentation,'' in \emph{Medical Imaging 2022: Image Processing}, vol. 12032.\hskip 1em plus 0.5em minus 0.4em\relax SPIE, 2022, pp. 792--799.

\bibitem{solar2009identifying}
M.~Solar, H.~Astudillo, G.~Valdes, M.~Iribarren, and G.~Concha, ``Identifying weaknesses for chilean e-government implementation in public agencies with maturity model,'' in \emph{Electronic Government: 8th International Conference, EGOV 2009, Linz, Austria, August 31-September 3, 2009. Proceedings 8}.\hskip 1em plus 0.5em minus 0.4em\relax Springer, 2009, pp. 151--162.

\bibitem{wisaeng2023u}
K.~Wisaeng, ``U-net++ dsm: improved u-net++ for brain tumor segmentation with deep supervision mechanism,'' \emph{IEEE Access}, vol.~11, pp. 132\,268--132\,285, 2023.

\bibitem{hatamizadeh2022unetr}
A.~Hatamizadeh, Y.~Tang, V.~Nath, D.~Yang, A.~Myronenko, B.~Landman, H.~R. Roth, and D.~Xu, ``Unetr: Transformers for 3d medical image segmentation,'' in \emph{Proceedings of the IEEE/CVF winter conference on applications of computer vision}, 2022, pp. 574--584.

\bibitem{biswas2023polyp}
R.~Biswas, ``Polyp-sam++: Can a text guided sam perform better for polyp segmentation?'' \emph{arXiv preprint arXiv:2308.06623}, 2023.

\bibitem{liu2022connecting}
R.~Liu, Z.~Lin, P.~Fu, Y.~Liu, and W.~Wang, ``Connecting targets via latent topics and contrastive learning: A unified framework for robust zero-shot and few-shot stance detection,'' in \emph{ICASSP 2022-2022 IEEE International Conference on Acoustics, Speech and Signal Processing (ICASSP)}.\hskip 1em plus 0.5em minus 0.4em\relax IEEE, 2022, pp. 7812--7816.

\bibitem{RN20}
X.~Liu, P.~Gao, T.~Yu, F.~Wang, and R.-Y. Yuan, ``Cswin-unet: Transformer unet with cross-shaped windows for medical image segmentation,'' \emph{Information Fusion}, vol. 113, p. 102634, 2025.

\bibitem{RN66}
L.~Huang, W.~Huang, H.~Gong, C.~Yu, and Z.~You, ``Pefnet: Position enhancement faster network for object detection in roadside perception system,'' \emph{IEEE Access}, 2023.

\bibitem{yu2019st}
B.~Yu, H.~Yin, and Z.~Zhu, ``St-unet: A spatio-temporal u-network for graph-structured time series modeling,'' \emph{arXiv preprint arXiv:1903.05631}, 2019.

\bibitem{RN67}
D.~Jha, N.~K. Tomar, V.~Sharma, and U.~Bagci, ``Transnetr: transformer-based residual network for polyp segmentation with multi-center out-of-distribution testing,'' in \emph{Medical Imaging with Deep Learning}.\hskip 1em plus 0.5em minus 0.4em\relax PMLR, 2024, pp. 1372--1384.

\bibitem{wang2022unetformer}
L.~Wang, R.~Li, C.~Zhang, S.~Fang, C.~Duan, X.~Meng, and P.~M. Atkinson, ``Unetformer: A unet-like transformer for efficient semantic segmentation of remote sensing urban scene imagery,'' \emph{ISPRS Journal of Photogrammetry and Remote Sensing}, vol. 190, pp. 196--214, 2022.

\bibitem{RN68}
D.~Jha, P.~H. Smedsrud, M.~A. Riegler, D.~Johansen, T.~De~Lange, P.~Halvorsen, and H.~D. Johansen, ``Resunet++: An advanced architecture for medical image segmentation,'' in \emph{2019 IEEE international symposium on multimedia (ISM)}.\hskip 1em plus 0.5em minus 0.4em\relax IEEE, 2019, pp. 225--2255.

\bibitem{reza2022contextual}
A.~Reza, H.~Moein, W.~Yuli, and M.~Dorit, ``Contextual attention network: Transformer meets u-net,'' \emph{arXiv preprint arXiv:2203.01932}, 2022.

\end{thebibliography}
\section{Acknowledgment}
Special thanks to Yuanyuan Li for her insightful comments and suggestions that helped improve this manuscript.

\EOD

\end{document}